\documentclass[10pt,twocolumn,letterpaper]{article}

\usepackage{cvpr}
\usepackage{times}
\usepackage{epsfig}
\usepackage{graphicx}
\usepackage{amsmath}
\usepackage{amssymb}
\usepackage{booktabs} 
\usepackage{algorithm}
\usepackage{algorithmic}
\usepackage{amsthm}
\usepackage{diagbox}
\usepackage{subfigure}
\usepackage{caption}
\usepackage{float}
\usepackage{epstopdf}
\usepackage{enumitem}

\usepackage{pifont}
\newcommand{\cmark}{\ding{51}}%
\newcommand{\xmark}{\ding{55}}%

\newtheorem{definition}{Definition}

\usepackage[breaklinks=true,bookmarks=false]{hyperref}

\cvprfinalcopy 

\def\cvprPaperID{520} 

\begin{document}

\title{Harmonizing Transferability and Discriminability for \\ Adapting Object Detectors}

\author{Chaoqi Chen$^1$, Zebiao Zheng$^1$, Xinghao Ding$^1$, Yue Huang$^{1}$\thanks{Corresponding author}, Qi Dou$^2$\\
 {$^1$~Fujian Key Laboratory of Sensing and Computing for Smart City,}\\
 {School of Informatics, Xiamen University, China}\\
 {$^2$~Department of Computer Science and Engineering, The Chinese University of Hong Kong}\\
 {\tt\small cqchen94@stu.xmu.edu.cn, zbzheng@stu.xmu.edu.cn}\\
 {\tt\small dxh@xmu.edu.cn, huangyue05@gmail.com, qdou@cse.cuhk.edu.hk}
}

\maketitle
\pagestyle{empty}
\thispagestyle{empty}

\begin{abstract}
Recent advances in adaptive object detection have achieved compelling results in virtue of adversarial feature adaptation to mitigate the distributional shifts along the detection pipeline. Whilst adversarial adaptation significantly enhances the transferability of feature representations, the feature discriminability of object detectors remains less investigated. Moreover, transferability and discriminability may come at a contradiction in adversarial adaptation given the complex combinations of objects and the differentiated scene layouts between domains. In this paper, we propose a Hierarchical Transferability Calibration Network (HTCN) that hierarchically (local-region/image/instance) calibrates the transferability of feature representations for harmonizing transferability and discriminability. The proposed model consists of three components: (1) Importance Weighted Adversarial Training with input Interpolation (IWAT-I), which strengthens the global discriminability by re-weighting the interpolated image-level features; (2) Context-aware Instance-Level Alignment (CILA) module, which enhances the local discriminability by capturing the underlying complementary effect between the instance-level feature and the global context information for the instance-level feature alignment; (3) local feature masks that calibrate the local transferability to provide semantic guidance for the following discriminative pattern alignment. Experimental results show that HTCN significantly outperforms the state-of-the-art methods on benchmark datasets.
\end{abstract}

\section{Introduction}
\label{intro}
Object detection has shown great success in the deep learning era, relying on representative features learned from large amount of labeled training data. Nevertheless, the object detectors trained on the source domain do not generalize well to a new target domain, due to the presence of domain shift~\cite{torralba2011unbiased}. This hinders the deployment of models in real-world situations where data distributions typically vary from one domain to another. Unsupervised Domain Adaptation~(UDA)~\cite{pan2010survey} serves as a promising solution to solve this problem by transferring knowledge from a labeled source domain to a fully unlabeled target domain.


A general practice in UDA is to bridge the domain gap by explicitly learning invariant representations between domains and achieving small error on the source domain,
which have achieved compelling performance on image classification~\cite{gong2012geodesic,tzeng2014deep,ganin2015unsupervised,sun2016deep,tzeng2017adversarial,saito2018maximum,Kang_2019_CVPR,Chen_2019_CVPR} and semantic segmentation~\cite{tsai2018learning,hoffman2018cycada,zou2018unsupervised,zhu2018penalizing,Li_2019_CVPR,lian2019constructing}. These UDA methods can fall into two main categories. The first category is statistics matching, which aims to match features across domains with statistical distribution divergence~\cite{gong2012geodesic,fernando2013unsupervised,long2015learning,long2017deep,zellinger2017central,peng2019moment}. The second category is adversarial learning, which aims to learn domain-invariant representations via domain adversarial training~\cite{ganin2015unsupervised,tzeng2017adversarial,shu2018dirt,long2018conditional,xie2018learning,Chen_2019_CVPR} or GAN-based pixel-level adaptation~\cite{bousmalis2017unsupervised,liu2017unsupervised,russo2018source,hu2018duplex,hoffman2018cycada}.

Regarding UDA for cross-domain object detection, several works~\cite{chen2018domain,saito2019strong,zhu2019adapting,cai2019exploring,kim2019diversify,He_2019_ICCV} have recently attempted to incorporate adversarial learning within de facto detection frameworks, e.g., Faster R-CNN~\cite{ren2015faster}.
With the local nature of detection tasks, current methods typically minimize the domain disparity at multiple levels via \emph{adversarial feature adaptation}, such as image and instance levels alignment~\cite{chen2018domain}, strong-local and weak-global alignment~\cite{saito2019strong}, local-region alignment based on region proposal~\cite{zhu2019adapting}, multi-level feature alignment with prediction-guided instance-level constraint~\cite{He_2019_ICCV}. They hold a common belief that harnessing adversarial adaptation helps yield appealing transferability.

However, transferability comes at a cost, \emph{i.e.,} adversarial adaptation would potentially impair the discriminability of target features since not all features are equally transferable.
Note that, in this paper, the
\emph{transferability} refers to the invariance of the learned representations across domains, and
\emph{discriminability} refers to the ability of the detector to localize and distinguish different instances.
Some recent studies~\cite{chen2019transferability,tsipras2019robustness} have also implied similar finding, but how to identify and calibrate the feature transferability still remains unclear.
This phenomenon would be more severe in cross-domain detection, given the complex combinations of various objects and the differentiated scene layouts between domains.
In other words, strictly aligning the entire feature distributions between domains by adversarial learning is prone to result in negative transfer, because \emph{the transferability of different levels (i.e., local-region, instance and image) is not explicitly elaborated in the object detector.}

In this work, we propose to harmonize transferability and discriminability for cross-domain object detection by developing a novel
Hierarchical Transferability Calibration Network (HTCN), which regularizes the adversarial adaptation by hierarchically calibrating the transferability of representations with improved discriminability.
Specifically, we first propose an Importance Weighted Adversarial Training with input Interpolation (IWAT-I) strategy, which aims to strengthen the global discriminability by re-weighting the interpolated feature space based on the motivation that \emph{not all samples are equally transferable especially after interpolation}.
Secondly, considering the structured scene layouts and the local nature of the detection task, we design a Context-aware Instance-Level Alignment (CILA) module to enhance the local discriminability by capturing \emph{the complementary effect between the instance-level feature and the global context information}. In particular, instead of simply concatenating these two terms, 
our approach resorts to the tensor product for more informative fusion.
Finally, upon observing that \emph{some local regions of the whole image are more descriptive and dominant than others}, we further enhance the local discriminability by proposing to compute local feature masks in both domains based on the shallow layer features for approximately guiding the semantic consistency in the following alignment, which can be seen as an attention-like module that capture the transferable regions in an unsupervised manner.

The proposed HTCN significantly extends the ability of previous adversarial-based adaptive detection methods by harmonizing the potential contradiction between transferability and discriminability. Extensive experiments show that the proposed method exceeds the state-of-the-art performance on several benchmark datasets for cross-domain detection. For example, we achieve 39.8\% mAP on adaptation from Cityscapes to Foggy-Cityscapes, outperforming the latest state-of-the-art adversarial-based adaptation methods~\cite{saito2019strong,zhu2019adapting,kim2019diversify,He_2019_ICCV} by a large margin (5.6\% on average) and approaching the upper bound (40.3\%). Our code is available at {\color{magenta}{https://chaoqichen.github.io/}}.
\section{Related Work}
\paragraph{Unsupervised Domain Adaptation}
Unsupervised domain adaptation (UDA) attempts to transfer knowledge from one domain to another by mitigating the distributional variations.
Recently, UDA has achieved extensive success, especially for image classification and semantic segmentation.
Typically, UDA methods propose to bridge different domains by matching the high-order statistics of source and target feature distributions in the latent space, such as Maximum Mean Discrepancy (MMD)~\cite{tzeng2014deep,long2015learning}, second-order moment~\cite{sun2016deep}, Central Moment Discrepancy (CMD)~\cite{zellinger2017central}, and Wasserstein distance~\cite{shen2018wasserstein}. With insights from the practice of Generative Adversarial Nets (GAN)~\cite{goodfellow2014generative}, tremendous works~\cite{ganin2016domain,tzeng2017adversarial,saito2018maximum,pei2018multi,xie2018learning,Chen_2019_CVPR} have been done by leveraging the two-player game to achieve domain confusion with Gradient Reversal Layer~(GRL) for feature alignment. In addition, other GAN-based works~\cite{bousmalis2017unsupervised,liu2017unsupervised,russo2018source,hu2018duplex,hoffman2018cycada,Tran_2019_CVPR} aim to achieve pixel-level adaptation in virtue of image-to-image translation techniques, \emph{e.g.,} CycleGAN~\cite{zhu2017unpaired}.
\vspace{-0.3cm}
\paragraph{UDA for Object Detection}
By contrast, there is relatively limited study on domain adaptation for object detection task, despite the impressive performance on single domain detection~\cite{ren2015faster,liu2016ssd,redmon2017yolo9000,lin2017focal,peng2018megdet}. Following the practice of conventional wisdom, Chen~\emph{et al.}~\cite{chen2018domain} pioneer this line of research, which propose a domain adaptive Faster R-CNN to reduce the distribution divergence in both image-level and instance-level by embedding adversarial feature adaptation into the two-stage detection pipeline. Saito~\emph{et al.}~\cite{saito2019strong} propose to align local receptive fields on shallow layers and image-level feature on deep layers, namely, strong local and weak global alignments. Similarly, He~\emph{et al.}~\cite{He_2019_ICCV} propose a hierarchical
domain feature alignment module and a weighted GRL to re-weight training samples. Zhu~\emph{et al.} and Cai~\emph{et al.}~\cite{zhu2019adapting,cai2019exploring} propose to exploit object proposal mining or object relations to achieve detailed local-region alignment in deep layers. Kim~\emph{et al.}~\cite{kim2019diversify} solve the adaptation problem from the perspective of domain diversification by randomly augmenting source and target domains into multiple domains, and then learning the invariant representations among domains. Nevertheless, all these UDA methods do not properly handle the the potential contradiction between transferability and discriminability when adapting object detectors in the context of adversarial adaptation.

\begin{figure*}[!t]
\centering
\includegraphics[width=0.95\textwidth]{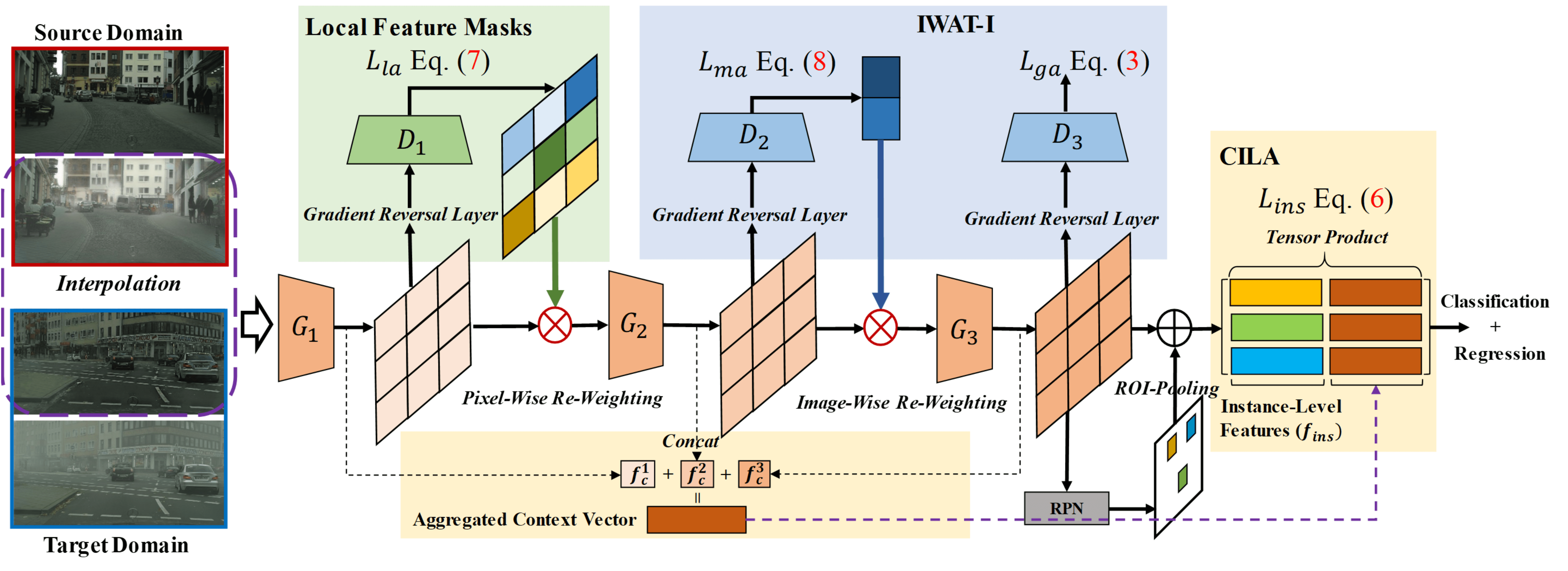}
\caption{The overall structure of the proposed HTCN. $D_1$ is \textbf{pixel-wise} domain discriminator, while $D_2$ and $D_3$ are \textbf{image-wise} domain discriminator. $G_1$, $G_2$, and $G_3$ denote the different level feature extractors. 
}\label{fig2}
\end{figure*}
\section{Hierarchical Transferability Calibration Network (HTCN)}
In this section, we present the technical details of the proposed method. The overall architecture of the proposed HTCN is shown in Fig.~\ref{fig2}, which consists of three modules, IWAT-I, CILA, and the local feature masks for semantic consistency. IWAT-I regularizes the image-level adversarial adaptation to calibrate the global transferability by re-weighting the interpolated feature space in image-wise. 
CILA regularizes the instance-level adversarial adaptation to calibrate the local transferability in virtue of tensor product to enable the informative interactions between the instance-level feature and the aggregated context vector.
\subsection{Problem Formulation}
For cross-domain object detection, it is required to simultaneously predict the bounding box locations and object categories. Formally, we have access to a labeled source dataset $\mathcal{D}_s=\{({x_i^s},{y_i^s},{b_i^s})\}_{i=1}^{N_s}$ (${y_i^s}\in{\mathcal{R}^{k\times{1}}}$, $b_i^s\in\mathcal{R}^{k\times{4}}$) of $N_s$ samples, and a target dataset $\mathcal{D}_t \! = \! \{{x_j^t}\}_{j=1}^{N_t}$ of $N_t$ unlabeled samples. The source and target domains share an identical label space, but violate the i.i.d. assumption as they are sampled from different data distributions. The goal of this paper is to learn an adaptive object detector, with the labeled $\mathcal{D}_s$ and unlabeled $\mathcal{D}_t$, which can perform well on the target domain. Following the mainstream cross-domain detection methods~\cite{chen2018domain,saito2019strong,zhu2019adapting,cai2019exploring,He_2019_ICCV}, the proposed HTCN is based on the Faster-RCNN~\cite{ren2015faster} framework.

As demonstrated in Section~\ref{intro}, transferability and discriminability may come at a contradiction in cross-domain detection tasks when using adversarial adaptation.
Motivated by this, our cross-domain detection approach resolves this problem from two perspectives: 1) calibrating the {\em transferability} by hierarchically identifying and matching the transferable local region features (Sec.~\ref{lfm}), holistic image-level features (Sec.~\ref{iwat}), and ROI-based instance-level features (Sec.~\ref{caila}), and 2) the hierarchical transferability-based cross-domain feature alignments, in turn, will improve the feature discriminability at multiple levels. 

\begin{figure}[!t]
\centering
\includegraphics[width=0.46\textwidth]{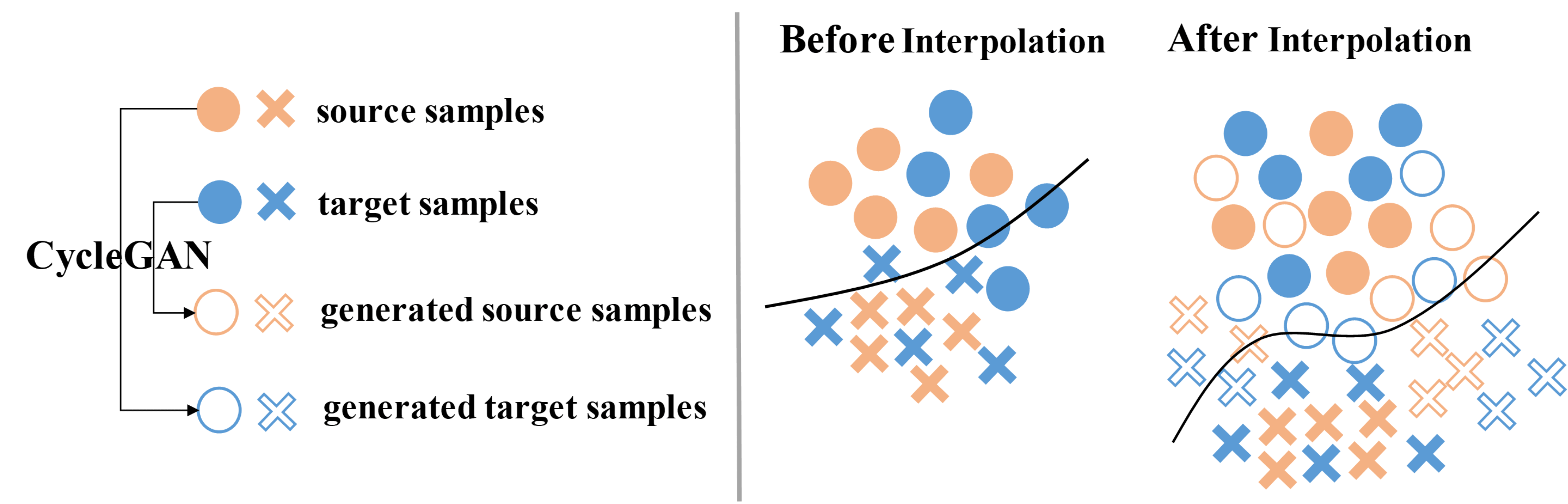}
\caption{Motivation of the interpolation operation for improving the source-biased decision boundary through generating synthetic samples from its counterpart domain to fill in the distributional gap between domains.
\vspace{-0.55cm}
}\label{fig3}
\end{figure}
\subsection{Importance Weighted Adversarial Training with Input Interpolation}
\label{iwat}
Domain adversarial training~\cite{ganin2015unsupervised} serves as a typical and powerful domain alignment approach to align feature distributions via a two-player game. Nevertheless, pure domain alignment may potentially deteriorate the semantic consistency and result in negative transfer, which has been extensively explored by numerous prior works~\cite{xie2018learning,zou2018unsupervised,long2018conditional,kumar2018co,Chen_2019_CVPR,Wang_2019_CVPR,Deng_2019_ICCV} in image classification and semantic segmentation tasks. By contrast, it is difficult or even impossible to explicitly encourage the cross-domain semantic consistency in object detection due to the distinct scene layouts, object co-occurrence, and background between domains. The representative semantic alignment strategies (\emph{e.g.,} prototype alignment~\cite{xie2018learning,Chen_2019_CVPR,Pan_2019_CVPR,Deng_2019_ICCV} or entropy regularization~\cite{zou2018unsupervised,shu2018dirt,Cicek_2019_ICCV}) would be no longer applicable.

To overcome the negative transfer in the context of cross-domain detection, the proposed IWAT-I adapts the source-biased decision boundary to target data through generating interpolation samples between domains, which implicitly induces the adversarial training to converge to a better saddle point and explicitly calibrate the global transferability to promote positive transfer.
The motivation of the interpolation based adversarial training is illustrated in Fig.~\ref{fig3}.
Without interpolation, the decision boundary learned by the adversarial training is prone to be source-biased, which will deteriorate its discriminability in the target domain.

The interpolation is implemented with CycleGAN~\cite{zhu2017unpaired} by generating synthetic samples from its counterpart domain to fill in the distributional gap between domains. 
Next, we aim to re-weight the interpolated data space based on their importance. The importance is associated with the cross-domain similarity, \emph{i.e.,} the higher the similarity is, the greater importance the sample is. Our key insight is that not all images are created equally in terms of transferability especially after interpolation. We aim to up-weight the most desirable samples while down-weight the irrelevant samples to calibrate the image-level transferability. 

Specifically, we leverage the uncertainty of the domain discriminator with respect to an input sample to discover transferable samples. The output of the discriminator $D_2$ \emph{w.r.t.} an input $x_i$ is $d_i \! = \! D_2(G_1 \circ G_2(x_i))$. Then, the uncertainty $v_i$ of each $x_i$ is measured by the information entropy \emph{w.r.t.} the output of the domain discriminator,
\begin{equation}\label{eq:gu}
v_i = H(d_i) = -d_i\cdot{\mathrm{log}(d_i)}-(1-d_i)\cdot{\mathrm{log}(1-d_i)}
\end{equation}
where $H(\cdot)$ is the entropy function. The weight of each image $x_i$ can then be computed as $1+v_i$. Images with high uncertainty (hard-to-distinguish by $D_2$) should be up-weighted, vice versa. The obtained uncertainty is then used to re-weight the feature representation as follows,
\begin{equation}\label{eq:rwf}
g_i=f_i\times(1+v_i)
\end{equation}
where $f_i$ is the feature before feeding into $D_2$. The input of $D_3$ is $G_3(g_i)$ and its adversarial loss is defined as,
\begin{equation}\label{eq:gal}
\mathcal{L}_{ga} = \mathbb{E}[\log(D_3(G_3(g_i^s))] + \mathbb{E}[1-\log(D_3(G_3(g_i^t))]
\end{equation}

\subsection{Context-Aware Instance-Level Alignment}
\label{caila}
Instance-level alignment refers to the ROI-Pooling based feature alignment, which has been explored by some prior efforts~\cite{chen2018domain,zhu2019adapting,He_2019_ICCV}. While these approaches are capable of alleviating the local instance deviations across domains (\emph{e.g.,} object scale, viewpoint, deformation, and appearance) to some extent, they may face a critical limitation that each feature vector of ROI layer represents the local object independently without considering the holistic \emph{context} information, which is an informative and decisive factor to the following detection and is prerequisite to induce accurate local instance alignment between domains. On the other hand, Yosinski~{\em et al.}~\cite{yosinski2014transferable} reveal that deep features must eventually transition from domain-agnostic to domain-specific along the network. Hence, the instance-level features obtained from deep layers may be distinct (discriminability) between domains. By contrast, the context vector is aggregated from the lower layer, which is relatively invariant (transferability) across domains. Thus, these two features can be complementary if we reasonably fuse them.

Motivated by the aforementioned findings, we propose a Context-aware Instance-Level Alignment~(CILA) loss that explicitly aligns the instance-level representations between domains based on the fusion of context vector and instance-wise representations. Formally, we denote the different levels of context vector as $\boldsymbol{f}_c^1$, $\boldsymbol{f}_c^2$, and $\boldsymbol{f}_c^3$ respectively. The instance-level features \emph{w.r.t.} the $j$-th region in the $i$-th image is denoted as $\boldsymbol{f}_{ins}^{i,j}$ and we omit the superscript for simplicity, $\boldsymbol{f}_{ins}$.
A simple approach for this fusion is concatenation, \emph{i.e.,} concatenating $\boldsymbol{f}_c^1$, $\boldsymbol{f}_c^2$, $\boldsymbol{f}_c^3$, and $\boldsymbol{f}_{ins}$ as a single vector $[\boldsymbol{f}_c^1, \boldsymbol{f}_c^2, \boldsymbol{f}_c^3, \boldsymbol{f}_{ins}]$. This aggregation strategy is extensively adopted by recent works~\cite{chen2018domain,saito2019strong,He_2019_ICCV} for regularizing the domain discriminator to achieve better adaptation. However, these approaches faces critical limitation. When using the concatenation strategy, the context features and the instance-level features are independent of each other, and thus they ignore the underlying complementary effect, which is crucial for a good domain adaptation. Moreover, these two features are asymmetric in our case, which impedes the using of some commonly used fusion methods, such as, element-wise product or averaging.

To overcome the aforementioned problems, we propose a non-linear fusion strategy with the following formulation,
\begin{equation}\label{fusion}
{\boldsymbol{f}}_{fus}=[\boldsymbol{f}_c^1, \boldsymbol{f}_c^2, \boldsymbol{f}_c^3]\otimes{\boldsymbol{f}_{ins}}
\end{equation}
where $\otimes$ denotes the tensor product operation and ${\boldsymbol{f}}_{fus}$ is the fused feature vector. By doing so, we are capable of producing informative interactions between the context feature and the instance-level feature. Such a non-linear strategy is beneficial for modeling some complex problems. However, this strategy still faces a dilemma of dimension explosion. Let us denote the aggregated context vector $[\boldsymbol{f}_c^1, \boldsymbol{f}_c^2, \boldsymbol{f}_c^3]$ as $\boldsymbol{f}_c$ and its dimension as $d_c$. Similarly, the dimension of $\boldsymbol{f}_{ins}$ is denoted as $d_{ins}$, and thus the dimension of ${\boldsymbol{f}}_{fus}$ will be $d_c\times{d_{ins}}$. In order to tackle the dimension explosion issue, we propose to leverage the randomized methods~\cite{long2018conditional,kar2012random} as an unbiased estimator of the tensor product. 
The final formulation is defined as follows,
\begin{equation}\label{schur}
\boldsymbol{f}_{fus}=\frac{1}{\sqrt{d}}(\boldsymbol{R}_1\boldsymbol{f}_c)\odot({\boldsymbol{R}_2\boldsymbol{f}_{ins}})
\end{equation}
where $\odot$ stands for the Hadamard product. $\boldsymbol{R}_1$ and $\boldsymbol{R}_2$ are random matrices and each of their element follows a symmetric distribution (\emph{e.g.,} Gaussian distribution and uniform distribution) with univariance. In our experiments, we follow the previous work~\cite{long2018conditional} by adopting the uniform distribution. $\boldsymbol{R}_1$ and $\boldsymbol{R}_2$ are sampled from uniform distribution only once and not updated during training. More details regarding Eq.~\ref{schur} are shown in our supplemental material.

Formally, the CA-ILA loss is defined as follows,
\begin{equation}\label{CA-ILA}
\small
\begin{split}
\mathcal{L}_{ins} &= -\frac{1}{N_s}\sum_{i=1}^{N_s}\sum_{i,j}\log(D_{ins}(\boldsymbol{f}_{fus}^{i,j})_s)\\
&= -\frac{1}{N_t}\sum_{i=1}^{N_t}\sum_{i,j}\log(1-D_{ins}(\boldsymbol{f}_{fus}^{i,j})_t)
\end{split}
\end{equation}

\subsection{Local Feature Mask for Semantic Consistency}
\label{lfm}
Although the scene layouts, object co-occurrence, and background may be distinct between domains, the description of the same object in different domains should be semantically invariant and can be matched, \emph{e.g.,} cars in different urban scenes should have similar sketch. Therefore, we assume that some local regions of the whole image are more descriptive and dominant than others. Motivated by this, we propose to compute local feature masks in both domains based on the shallow layer features for approximately guiding the semantic consistency in the following adaptation, which can be seen as an attention-like module that capture the transferable regions in an unsupervised manner. 

Technically, the feature masks $m_f^s$ and $m_f^t$ are computed by utilizing the uncertainty of the local domain discriminator $D_1$. $D_1$ is a pixel-wise discriminator. Suppose that the feature maps from $G_1$ have width of $W$ and height of $H$. Therefore, the pixel-wise adversarial training loss $\mathcal{L}_{la}$ is formulated as follows,
\begin{small}
\begin{equation}\label{eq:localadv}
\begin{split}
\mathcal{L}_{la}&=\frac{1}{N_s \! \cdot \! HW}\sum_{i=1}^{N_s}\sum_{k=1}^{HW}\log(D_1(G_1(x_i^s)_k))^2 \\
&+\frac{1}{N_t \! \cdot \! HW}\sum_{i=1}^{N_t}\sum_{k=1}^{HW}\log(1-D_1(G_1(x_i^t)_k))^2,
\end{split}
\end{equation}
\end{small}
where $(G_1(x_i))_k$ denotes the feature vector of the $k$th location in the feature map obtained from $G_1(x_i)$. For ease of denotation, we omit the superscript from $x_i^s$ and $x_i^t$ as $x_i$, when it applies. Hereafter, $(G_1(x_i))_k$ is denoted as $r_i^k$. Note that a location in the abstracted feature map corresponds to a region in the original image with a certain receptive field.
For each region, the output of discriminator $D_1$ is represented by $d_i^k = D_1(r_i^k)$. Similar to Eq.~\eqref{eq:gu}, the uncertainty from $D_1$ at each region is computed as $v(r_i^k)=H(d_i^k)$.
Based on the computed uncertainty map, the feature mask of each region $m_f^k$ is further defined as $m_f^k=2-v(r_i^k)$, \emph{i.e.,} the less uncertainty regions are more transferable. To this end, to incorporate the local feature masks into the detection pipeline, we re-weight the local features by $\tilde{r}_i^k \gets r_i^k \cdot m_i^k$.
In that way, the informative regions will be assigned a higher weight, while other less informative regions will be suppressed. The source and target feature masks are computed respectively to semantically guide the following high-level feature adaptation. To this end, the adversarial loss of $D_2$ is defined as follows,
\begin{equation}\label{eq:mal}
\small
\mathcal{L}_{ma} = \mathbb{E}[\log(D_2(G_2(\hat{f}_i^s))] + \mathbb{E}[1-\log(D_2(G_2(\hat{f}_i^t))]
\end{equation}
where $\hat{f}_i^s$ and $\hat{f}_i^t$ denote the whole pixel-wise re-weighted feature maps.
\subsection{Training Loss}
The detection loss includes $\mathcal{L}_{\text{cls}}$ and $\mathcal{L}_{\text{reg}}$ which measure how accurate of the classification, and the overlap of the predicted and ground-truth bounding boxes. Combining all the presented parts, the overall objective function for the proposed model is,
\begin{equation}\label{eq:all}
\footnotesize
\max\limits_{D_1,D_2,D_3}\;\min\limits_{G_1,G_2,G_3} \mathcal{L}_{\text{cls}} + \mathcal{L}_{\text{reg}}-\lambda(\mathcal{L}_{la}+\mathcal{L}_{ma}+\mathcal{L}_{ga}+\mathcal{L}_{ins}),
\end{equation}
where $\lambda$ is parameters balancing loss components.
\subsection{Theoretical Insights}
We provide theoretical insights of our approach \emph{w.r.t.} the domain adaptation theory. We assume that the cross-domain detection by unconstrained adversarial training can be seen as a non-conservative domain adaptation~\cite{ben2010impossibility,shu2018dirt} problem due to the potential contradiction between transferability and discriminability. Conservative domain adaptation~\cite{ben2010impossibility} refers to a scenario that a learner only need to find the optimal hypothesis regarding the labeled source samples and evaluate the performance of this hypothesis on the target domain by using the unlabeled target samples.
\begin{definition}\label{ncda}
Let $\mathcal{H}$ be the hypothesis class. Given two different domains $\mathcal{S}$, $\mathcal{T}$, in non-conservative domain adaptation, we have the following inequality,
\begin{equation}\label{gap}
\small
\begin{split}
R_{\mathcal{T}}(h^t)&<R_{\mathcal{T}}(h^*), {\rm{where}} \\
h^*&=\arg\min\limits_{{h}\in\mathcal{H}}\;R_{\mathcal{S}}(h)+R_{\mathcal{T}}(h), \\
h^t&=\arg\min\limits_{{h}\in\mathcal{H}}\;R_{\mathcal{T}}(h)
\end{split}
\end{equation}
where $R_{\mathcal{S}}(\cdot)$ and $R_{\mathcal{T}}(\cdot)$ denote the expected risk on source and target domains.
\end{definition}

Def.~\ref{ncda} shows that there exists an optimality gap between the optimal source detector and the optimal target detector in non-conservative domain adaptation, which results from the contradiction between transferability and discriminability. Strictly matching the whole feature distributions between domains (\emph{i.e.,} aiming to find a hypothesis that simultaneously minimizes the source and target expected errors) \emph{inevitably} results in sub-optimal solution according to Def.~\ref{ncda}. Hence, we are required to design a model that promotes the parts of transferable features and alleviates those non-transferable features. Theoretically, our work is not to explicitly seek $h^t$ in the target domain due to the absence of ground-truth labels, but to solve the non-conservative domain adaptation problem and minimize the upper bound of the expected target error, \emph{i.e.,} $R_{\mathcal{T}}(h)$.

The theory of domain adaptation~\cite{ben2010theory} bounds the expected error on the target domain as follows,
\begin{equation}\label{eq:bound}
\small
\forall{h}\in{\mathcal{H}},R_{\mathcal{T}}(h)\leq{R_{\mathcal{S}}(h)}+\frac{1}{2}d_{\mathcal{H}\Delta\mathcal{H}}(\mathcal{S},\mathcal{T})+C
\end{equation}
where $R_{\mathcal{S}}$ denotes the expected error on the source domain, $d_{\mathcal{H}\Delta\mathcal{H}}(\mathcal{S},\mathcal{T})$ stands for the the domain divergence and associated with the feature transferability, and $C$ is the error of the ideal joint hypothesis (\emph{i.e.,} $h^*$ in Eq.~\eqref{gap}) and associated with the feature discriminability. In Inequality~\eqref{eq:bound}, $R_{\mathcal{S}}$ can be easily minimized by a deep network since we have source labels. More importantly, our approach hierarchically identify the transferable region/image/instance and enhance their transferability to minimize $d_{\mathcal{H}\Delta\mathcal{H}}(\mathcal{S},\mathcal{T})$ by the local feature masks, IWAT-I, and CILA. And we improve the discriminability to minimize $C$ by the hierarchical transferability-based cross-domain feature alignments. By doing so, we are able to mitigate the contradiction between transferability and discriminability.
\vspace{-0.1cm}
\section{Experiments}
\subsection{Datasets}
\paragraph{Cityscapes~$\rightarrow$~Foggy-Cityscapes.} \textbf{Cityscapes}~\cite{cordts2016cityscapes} is collected from the street scenarios of different cities. It includes 2, 975 images in the training set and 500 images in the testing set. We used the training set during training and evaluated on the testing set by following~\cite{saito2019strong}. The images are captured by a car-mounted video camera in normal weather conditions. Joining previous practices~\cite{chen2018domain,saito2019strong}, we utilize the rectangle of instance mask to obtain bounding boxes for our experiments. \textbf{Foggy-Cityscapes}~\cite{cordts2016cityscapes} are rendered from Cityscape by using depth information to simulate the foggy scenes. The bounding box annotations are inherited from the Cityscapes dataset. Note that we utilize the training set of Foggy-Cityscapes as the target domain.
\vspace{-0.5cm}
\paragraph{PASCAL~$\rightarrow$~Clipart.} We use the combination of the training and validation set in \textbf{PASCAL}~\cite{everingham2010pascal} as the source domain by following~\cite{saito2019strong}. \textbf{Clipart} is from the Watercolor datasets~\cite{inoue2018cross} and used as the target domain, which contains 1K images and have the same 20 categories as PASCAL.
\vspace{-0.5cm}
\paragraph{Sim10K~$\rightarrow$~Cityscapes.} \textbf{Sim10K}~\cite{johnson2017driving} is a dataset produced based on the computer game Grand Theft Auto V (GTA V). It contains 10,000 images of the synthetic driving scene with 58,071 bounding boxes of the car. All images of Sim10K are utilized as the source domain.

\subsection{Implementation Details}
The detection model follows the setting in~\cite{chen2018domain,zhu2019adapting,saito2019strong} that adopt Faster-RCNN~\cite{ren2015faster} with VGG-16~\cite{simonyan2014very} or ResNet-101~\cite{he2016deep} architectures. The parameters of VGG-16 and ResNet-101 are fine-tuned from the model pre-trained on ImageNet. In experiments, the shorter side of each input image is resized to 600. At each iteration, we input one source image and one target-like source image as the source domain, while the target domain includes one target image and one source-like target image. We evaluate the adaptation performance by reporting mean average precision (mAP) with a IoU threshold of 0.5. We utilize stochastic gradient descent (SGD) for the training procedure with a momentum of 0.9 and the initial learning rate is set to 0.001, which is decreased to 0.0001 after 50K iterations. For Cityscapes~$\rightarrow$~Foggy-Cityscapes and PASCAL~$\rightarrow$~Clipart, we set $\lambda=1$ in Eq.~\eqref{eq:all}. For Sim10K~$\rightarrow$~Cityscapes, we set $\lambda=0.1$. The hyper-parameters of the detection model are set by following~\cite{ren2015faster}. All experiments are implemented by the Pytorch framework.

\begin{center}
\begin{table*}[thb]
\caption{Results on adaptation from Cityscapes to Foggy-Cityscapes. Average precision (\%) is reported on the target domain. Note that the backbone of MTOR is ResNet-50, while the others are VGG-16.}\label{table1}
\centering
\small
\qquad 
\setlength\tabcolsep{1.5pt}
\scalebox{0.9}{
\begin{tabular*}{\hsize}{@{}@{\extracolsep{\fill}}c|ccccccccc@{}}
\toprule
Methods & Person & Rider & Car & Truck & Bus & Train & Motorbike & Bicycle & mAP \\
\hline
Source Only~\cite{ren2015faster} & 24.1 & 33.1 & 34.3 & 4.1 & 22.3 & 3.0 & 15.3 & 26.5 & 20.3 \\
DA-Faster~(CVPR'18)~\cite{chen2018domain} & 25.0 & 31.0 & 40.5 & 22.1 & 35.3 & 20.2 & 20.0 & 27.1 & 27.6 \\
SCDA~(CVPR'19)~\cite{zhu2019adapting} & \textbf{33.5} & 38.0 & \textbf{48.5} & 26.5 & 39.0 & 23.3 & 28.0 & 33.6 & 33.8 \\
MAF~(ICCV'19)~\cite{He_2019_ICCV} & 28.2 & 39.5 & 43.9 & 23.8 & 39.9 & 33.3 & 29.2 & 33.9 & 34.0 \\
SWDA~(CVPR'19)~\cite{saito2019strong} & 29.9 & 42.3 & 43.5 & 24.5 & 36.2 & 32.6 & 30.0 & 35.3 & 34.3 \\
DD-MRL~(CVPR'19)~\cite{kim2019diversify} & 30.8 & 40.5 & 44.3 & 27.2 & 38.4 & 34.5 & 28.4 & 32.2 & 34.6 \\
MTOR*~(CVPR'19)~\cite{cai2019exploring} & 30.6 & 41.4 & 44.0 & 21.9 & 38.6 & 40.6 & 28.3 & 35.6 & 35.1 \\
\hline
HTCN & 33.2 & \textbf{47.5} & 47.9 & \textbf{31.6} & \textbf{47.4} & \textbf{40.9} & \textbf{32.3} & \textbf{37.1} & \textbf{39.8} \\
\hline
Upper Bound & 33.2 & 45.9 & 49.7 & 35.6 & 50.0 & 37.4 & 34.7 & 36.2 & 40.3 \\
\bottomrule
\end{tabular*}}
\end{table*}
\end{center}

\begin{center}
\begin{table*}[htb]
\caption{Results on adaptation from PASCAL VOC to Clipart Dataset (\%). The results of SWDA*~(only G) are cited from~\cite{saito2019strong}, which only uses the global alignment. The backbone network is ResNet-101.}\label{table2}
\vspace{-0.2cm}
\centering
\footnotesize
\qquad 
\setlength\tabcolsep{1.5pt}
\begin{tabular*}{\hsize}{@{}@{\extracolsep{\fill}}c|ccccccccccccccccccccc@{}}
\toprule
Methods & aero & bcycle & bird & boat & bottle & bus & car & cat & chair & cow & table & dog & hrs & bike & prsn & plnt & sheep & sofa & train & tv & mAP \\
\hline
Source Only~\cite{ren2015faster} & \textbf{35.6} & 52.5 & 24.3 & 23.0 & 20.0 & 43.9 & 32.8 & 10.7 & 30.6 & 11.7 & 13.8 & 6.0 & 36.8 & 45.9 & 48.7 & 41.9 & 16.5 & 7.3 & 22.9 & 32.0 & 27.8 \\
DA-Faster~\cite{chen2018domain} & 15.0 & 34.6 & 12.4 & 11.9 & 19.8 & 21.1 & 23.2 & 3.1 & 22.1 & 26.3 & 10.6 & 10.0 & 19.6 & 39.4 & 34.6 & 29.3 & 1.0 & 17.1 & 19.7 & 24.8 & 19.8 \\
WST-BSR~\cite{kim2019self} & 28.0 & \textbf{64.5} & 23.9 & 19.0 & 21.9 & \textbf{64.3} & \textbf{43.5} & 16.4 & \textbf{42.2} & 25.9 & \textbf{30.5} & 7.9 & 25.5 & 67.6 & 54.5 & 36.4 & 10.3 & \textbf{31.2} & \textbf{57.4} & 43.5 & 35.7 \\
SWDA*~(only G)~\cite{saito2019strong} & 30.5 & 48.5 & 33.6 & 24.8 & 41.2 & 48.9 & 32.4 & 17.2 & 34.5 & 55.0 & 19.0 & 13.6 & 35.1 & 66.2 & \textbf{63.0} & 45.3 & 12.5 & 22.6 & 45.0 & 38.9 & 36.4 \\
SWDA~\cite{saito2019strong} & 26.2 & 48.5 & 32.6 & \textbf{33.7} & 38.5 & 54.3 & 37.1 & \textbf{18.6} & 34.8 & \textbf{58.3} & 17.0 & 12.5 & 33.8 & 65.5 & 61.6 & \textbf{52.0} & 9.3 & 24.9 & 54.1 & 49.1 & 38.1 \\
\hline
HTCN & 33.6 & 58.9 & \textbf{34.0} & 23.4 & \textbf{45.6} & 57.0 & 39.8 & 12.0 & 39.7 & 51.3 & 21.1 & \textbf{20.1} & \textbf{39.1} & \textbf{72.8} & \textbf{63.0} & 43.1 & \textbf{19.3} & 30.1 & 50.2 & \textbf{51.8} & \textbf{40.3} \\
\bottomrule
\end{tabular*}
\end{table*}
\end{center}
\vspace{-1.8cm}
\subsection{Comparisons with State-of-the-Arts}
\paragraph{State-of-the-arts.} We compare our approach with various state-of-the-art cross-domain detection methods, including Domain adaptive Faster-RCNN (\textbf{DA-Faster})~\cite{chen2018domain}, Selective Cross-Domain Alignment~(\textbf{SCDA})~\cite{zhu2019adapting}, Multi-Adversarial Faster-RCNN (\textbf{MAF})~\cite{He_2019_ICCV}, Strong-Weak Distribution Alignment~(\textbf{SWDA})~\cite{saito2019strong}, Domain Diversification and Multi-domain-invariant Representation Learning (\textbf{DD-MRL})~\cite{kim2019diversify}, and Mean Teacher with Object Relations (\textbf{MTOR})~\cite{cai2019exploring}. For all above-mentioned methods, we cite the result in their original papers respectively.

Table~\ref{table1} shows the results of adaptation from Cityscapes to Foggy-Cityscapes. Source only stands for the model that is trained only using source images without adaptation. The proposed HTCN significantly outperforms all comparison methods and improves over state-of-the-art results by +4.7\% on average (from 35.1\% to 39.8\%), which is very close to the upper bound of this transfer task (only 0.5\% apart). The compelling results clearly demonstrate that HTCN can learn more discriminative representation in the target domain by calibrating the transferability of the feature representations.

The results of PASCAL~$\rightarrow$~Clipart are reported in Table~\ref{table2}. HTCN achieves state of the art mAP on adaptation between two dissimilar domains (from real images to artistic images)~\cite{saito2019strong}, which clearly verifies the robustness of HTCN on the challenging scenario.

Table~\ref{table3} shows the results of adaptation from SIM10K (synthetic) to Cityscapes (real). Our HTCN outperforms all comparison methods, which further verifies the effectiveness of our hierarchical transferability calibration insights.
\subsection{Further Empirical Analysis}
\paragraph{Ablation Study.} We conduct the ablation study by evaluating variants of HTCN. The results are reported in Table~\ref{table4}. We can see that all the proposed modules are designed reasonably and when any one of these modules is removed, the performance drops accordingly. Note that \textbf{HTCN-w/o Context Information} denotes that we remove the context information from the CILA module but the instance-level alignment is preserved. \textbf{HTCN-w/o Tensor Product} denotes that we utilize the vanilla concatenation to replace the tensor product for fusing context and instance-level features. \textbf{HTCN (full)} denotes the full model.
\vspace{-0.2cm}
\begin{center}
\begin{table}[htb]
\caption{Results on Sim10K~$\rightarrow$~Cityscapes (\%). L, G, LFM, CI indicate local region alignment, global image alignment, local feature mask, and context-vector based instance-level alignment. The backbone network is VGG-16.}\label{table3}
\vspace{-0.2cm}
\centering
\setlength{\belowcaptionskip}{-0.5cm}
\small
\qquad 
\setlength\tabcolsep{1.5pt}
\scalebox{0.91}{
\begin{tabular*}{\hsize}{@{}@{\extracolsep{\fill}}cccccc@{}}
\toprule
Methods & L & G & LFM & CI & AP on \emph{car} \\
\hline
Source Only~\cite{ren2015faster} & \xmark & \xmark & \xmark & \xmark & 34.6 \\
DA-Faster~\cite{chen2018domain} & \cmark & \cmark & \xmark & \xmark & 38.9 \\
SWDA~\cite{saito2019strong} & \cmark & \cmark & \xmark & \xmark & 40.1 \\
MAF~\cite{He_2019_ICCV} & \cmark & \cmark & \xmark & \xmark & 41.1 \\
\hline
HTCN & \cmark & \cmark & \cmark & \cmark & \textbf{42.5} \\
\bottomrule
\end{tabular*}}
\end{table}
\end{center}

\begin{figure}[!t]
\centering
\setlength{\belowcaptionskip}{-0.5cm}
\includegraphics[width=0.4\textwidth]{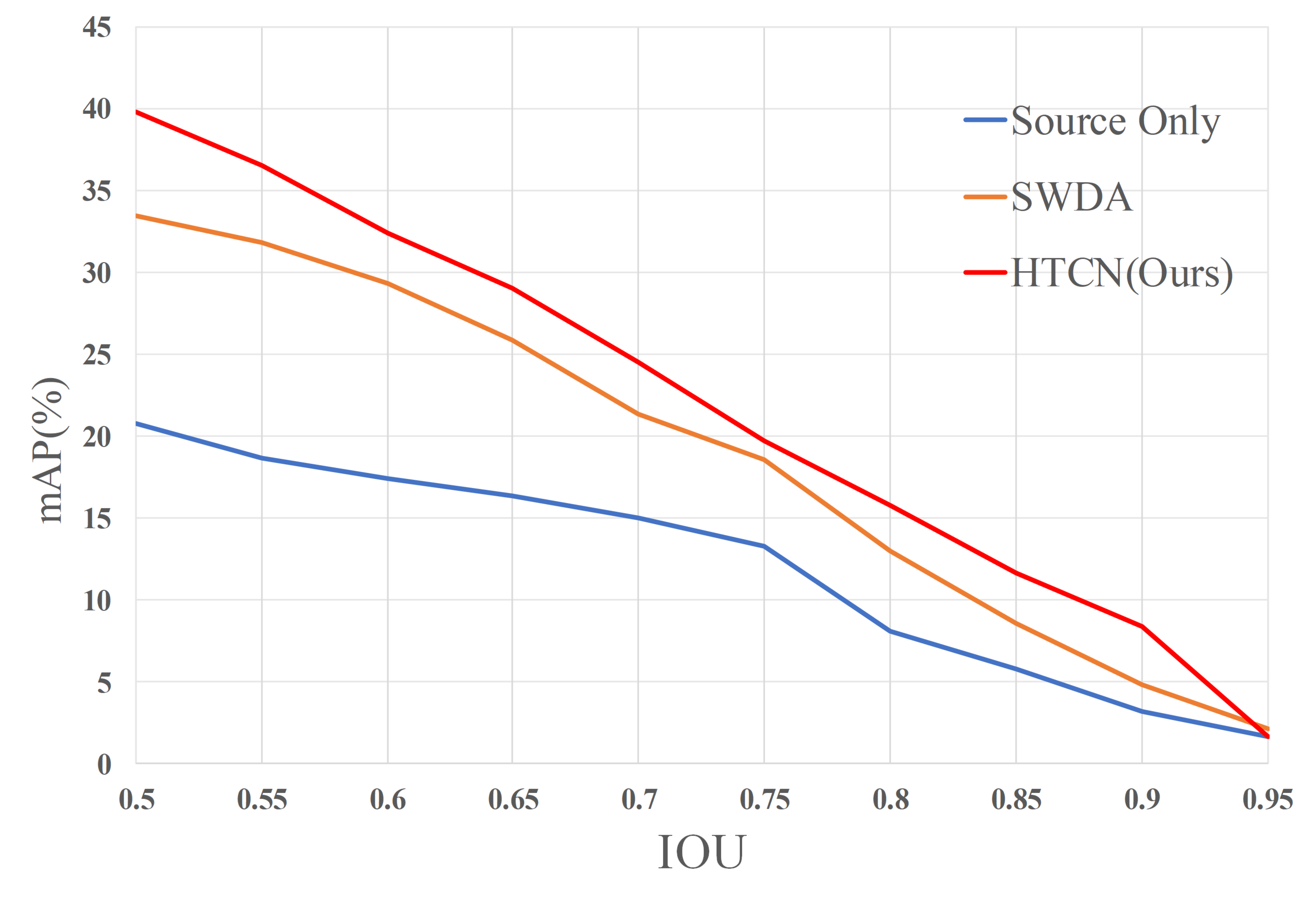}
\vspace{-0.2cm}
\caption{The performance with the variation of IOU thresholds on transfer task Cityscapes~$\rightarrow$~Foggy-Cityscapes.}\label{fig6}
\end{figure}

\vspace{-1.3cm}
\paragraph{Influence of IOU thresholds.} Figure~\ref{fig6} shows the performance of different models (\emph{i.e.,} Source Only, SWDA~\cite{saito2019strong}, and HTCN) with the variation of IOU thresholds. We found that the mAP continuously drops with the increasing of the IOU threshold and close to zero in the end. It is noteworthy that the proposed HTCN significantly outperforms the comparison methods on the IOU range 0.5-0.9, which implies that our HTCN can provide more accurate and robust bounding boxes regression.
\vspace{-0.2cm}

\begin{center}
\begin{table*}[!t]
\caption{Ablation of HTCN on Cityscapes~$\rightarrow$~Foggy-Cityscapes.}\label{table4}
\centering
\vspace{-0.2cm}
\small
\scalebox{0.82}{
\begin{tabular*}{\hsize}{@{}@{\extracolsep{\fill}}c|ccccccccc@{}}
\toprule
Methods & Person & Rider & Car & Truck & Bus & Train & Motorbike & Bicycle & mAP \\
\hline
Source Only & 24.1 & 33.1 & 34.3 & 4.1 & 22.3 & 3.0 & 15.3 & 26.5 & 20.3 \\
HTCN-w/o IWAT-I & 30.5 & 42.0 & 44.3 & 21.6 & 39.4 & 34.1 & 32.3 & 33.0 & 34.7 \\
HTCN-w/o CILA & 32.9 & 45.9 & \textbf{48.5} & 27.6 & 44.6 & 22.1 & \textbf{34.1} & 37.6 & 36.6 \\
HTCN-w/o Local Feature Masks & 32.9 & 46.2 & 48.2 & 31.1 & 47.3 & 33.3 & 33.0 & \textbf{39.0} & 38.9 \\
HTCN-w/o Interpolation & 32.8 & 45.6 & 44.8 & 26.5 & 44.3 & 36.9 & 32.0 & 37.1 & 37.5 \\
HTCN-w/o Context Information &  30.0 & 43.0 & 44.4 & 28.2 & 43.1 & 32.3 & 28.7 & 33.7 & 35.4 \\
HTCN-w/o Tensor Product & \textbf{33.3} & 46.7 & 47.6 & 28.6 & 46.1 & 36.4 & 32.6 & 37.2 & 38.6 \\
\hline
HTCN (full) & 33.2 & \textbf{47.5} & 47.9 & \textbf{31.6} & \textbf{47.4} & \textbf{40.9} & 32.3 & 37.1 & \textbf{39.8} \\
\bottomrule
\end{tabular*}}
\end{table*}
\end{center}

\begin{figure*}[!t]
\centering
\includegraphics[width=0.85\textwidth]{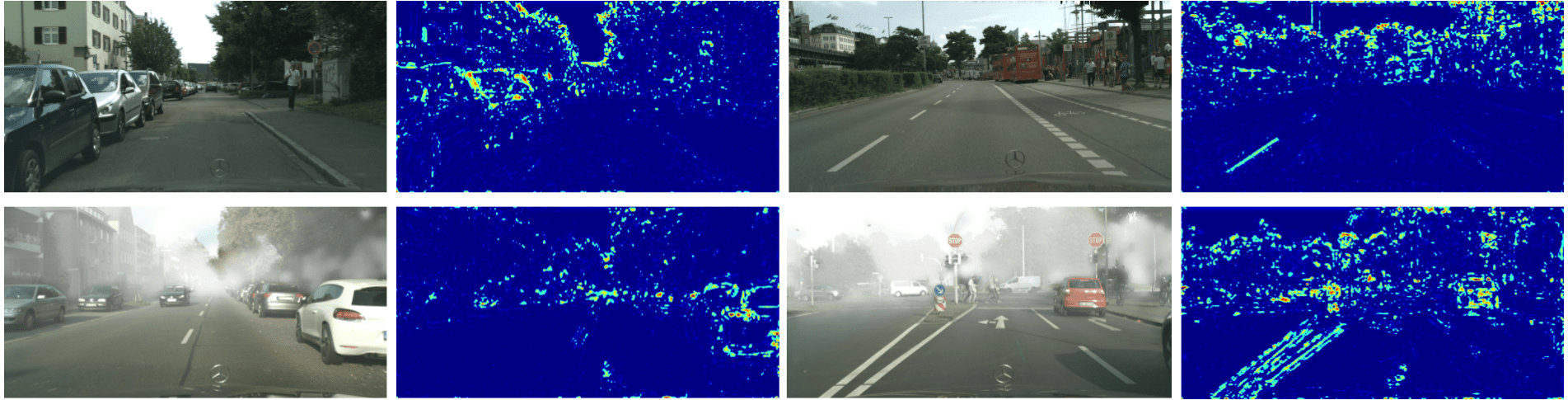}
\caption{Illustration of the local feature masks on adaptation task Cityscapes~(Top)~$\rightarrow$~Foggy-Cityscapes~(Bottom).}\label{fig5}
\end{figure*}

\begin{figure*}[!t]
\centering
\setlength{\belowcaptionskip}{-0.5cm}
\includegraphics[width=0.85\textwidth]{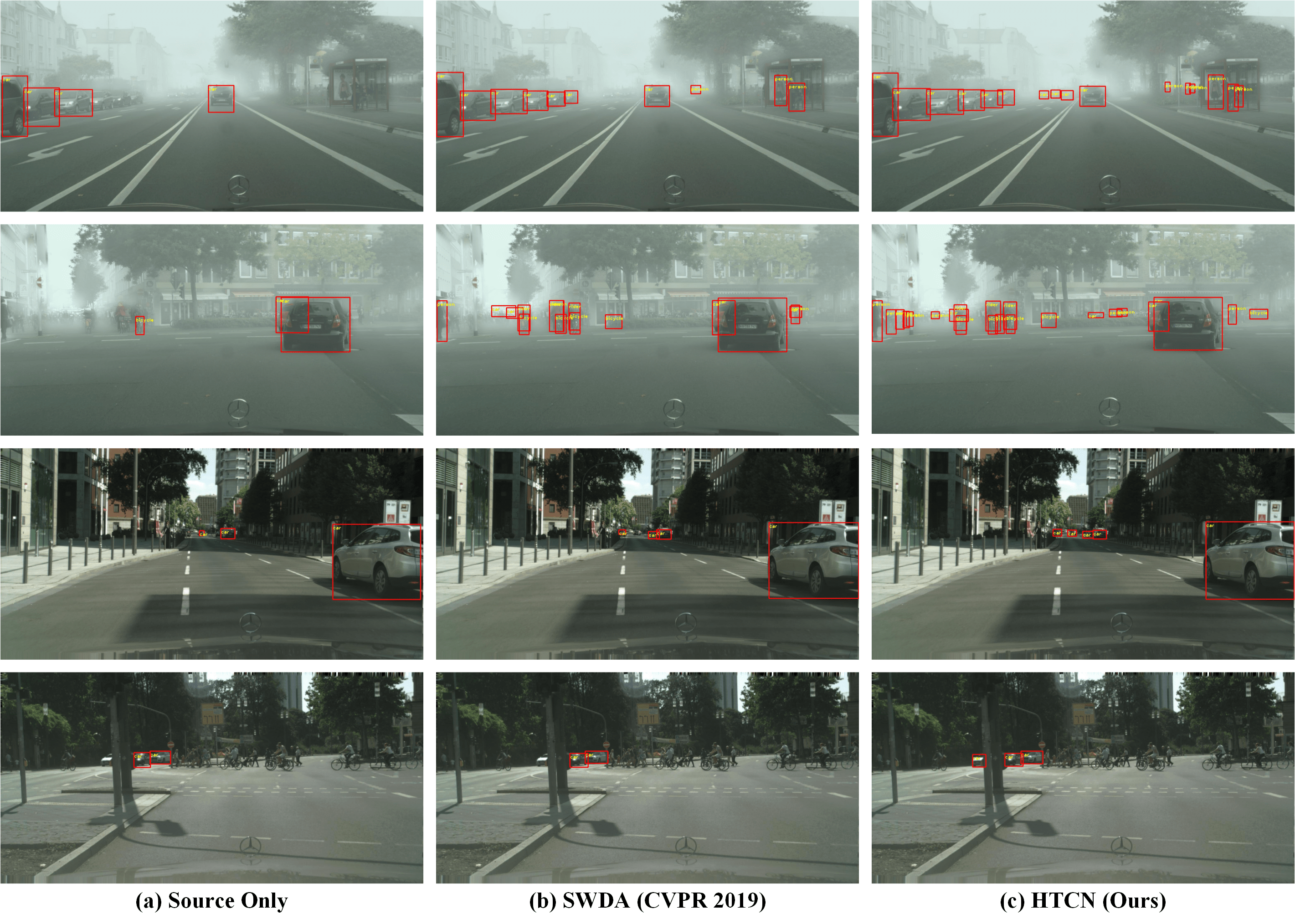}
\caption{Illustration of the detection results on the target domain. 
First and second rows: Cityscapes~$\rightarrow$~Foggy-Cityscapes. Third and fourth rows: Sim10K$\rightarrow$Cityscapes. Best view in color.}\label{fig4}
\end{figure*}
\vspace{-0.5cm}
\paragraph{Visualization of Local Feature Masks.} Figure~\ref{fig5} visualizes the proposed local feature masks on source and target domains. The brighter the color is, the larger the weight value is. We can observe that the source and target feature masks demonstrate an edge-aware pattern, which focuses on the edge of different instances (\emph{e.g.} car and person) and some other descriptive regions (\emph{e.g.} building and traffic sign). Due to the presence of the large distributional variations between domains, forcefully matching the two domains is prone to result in negative transfer since not all regions are informative and transferable. By contrast, the local feature masks make the adaptation network up-weight the semantically descriptive and informative regions to yield better discriminability by transferability calibration.
\vspace{-0.5cm}
\paragraph{Example of Detection Results.} Figure~\ref{fig4} illustrates the example of detection results on transfer tasks Cityscapes~$\rightarrow$~Foggy-Cityscapes and Sim10K$\rightarrow$Cityscapes, respectively. The proposed HTCN consistently outperforms both Source Only~\cite{ren2015faster} and SWDA~\cite{saito2019strong} models in different tasks. For example, in the results of Foggy-Cityscapes, HTCN is capable of detecting those obscured instances with accurate bounding box predictions.

\section{Conclusion}
This paper presented a novel framework called Hierarchical Transferability Calibration Network to harmonize transferability and discriminability in the context of adversarial adaptation for adapting object detectors by exploring the transferability of different local-regions, images, and instances. The extensive experiments demonstrate that our approach yields state-of-the-art performance for adapting object detectors on several benchmark datasets.

\section{Acknowledgements}
The work is supported in part by National Natural Science Foundation of China under Grants 81671766, 61571382, U19B2031, 61971369, U1605252, 81671674, in part of Fundamental Research Funds for the Central Universities 20720180059 and 20720190116 and in part of CCF-Tencent open fund.

{\small
\bibliographystyle{ieee_fullname}
\bibliography{egpaper_final}

\begin{thebibliography}{10}\itemsep=-1pt

\bibitem{ben2010theory}
Shai Ben-David, John Blitzer, Koby Crammer, Alex Kulesza, Fernando Pereira, and
  Jennifer~Wortman Vaughan.
\newblock A theory of learning from different domains.
\newblock {\em Machine learning}, 79(1-2):151--175, 2010.

\bibitem{ben2010impossibility}
Shai Ben-David, Tyler Lu, Teresa Luu, and D{\'a}vid P{\'a}l.
\newblock Impossibility theorems for domain adaptation.
\newblock In {\em International Conference on Artificial Intelligence and
  Statistics}, pages 129--136, 2010.

\bibitem{bousmalis2017unsupervised}
Konstantinos Bousmalis, Nathan Silberman, David Dohan, Dumitru Erhan, and Dilip
  Krishnan.
\newblock Unsupervised pixel-level domain adaptation with generative
  adversarial networks.
\newblock In {\em CVPR}, 2017.

\bibitem{cai2019exploring}
Qi Cai, Yingwei Pan, Chong-Wah Ngo, Xinmei Tian, Lingyu Duan, and Ting Yao.
\newblock Exploring object relation in mean teacher for cross-domain detection.
\newblock In {\em CVPR}, pages 11457--11466, 2019.

\bibitem{Chen_2019_CVPR}
Chaoqi Chen, Weiping Xie, Wenbing Huang, Yu Rong, Xinghao Ding, Yue Huang,
  Tingyang Xu, and Junzhou Huang.
\newblock Progressive feature alignment for unsupervised domain adaptation.
\newblock In {\em CVPR}, pages 627--636, 2019.

\bibitem{chen2019transferability}
Xinyang Chen, Sinan Wang, Mingsheng Long, and Jianmin Wang.
\newblock Transferability vs. discriminability: Batch spectral penalization for
  adversarial domain adaptation.
\newblock In {\em ICML}, pages 1081--1090, 2019.

\bibitem{chen2018domain}
Yuhua Chen, Wen Li, Christos Sakaridis, Dengxin Dai, and Luc Van~Gool.
\newblock Domain adaptive faster r-cnn for object detection in the wild.
\newblock In {\em CVPR}, pages 3339--3348, 2018.

\bibitem{Cicek_2019_ICCV}
Safa Cicek and Stefano Soatto.
\newblock Unsupervised domain adaptation via regularized conditional alignment.
\newblock In {\em ICCV}, 2019.

\bibitem{cordts2016cityscapes}
Marius Cordts, Mohamed Omran, Sebastian Ramos, Timo Rehfeld, Markus Enzweiler,
  Rodrigo Benenson, Uwe Franke, Stefan Roth, and Bernt Schiele.
\newblock The cityscapes dataset for semantic urban scene understanding.
\newblock In {\em CVPR}, pages 3213--3223, 2016.

\bibitem{Deng_2019_ICCV}
Zhijie Deng, Yucen Luo, and Jun Zhu.
\newblock Cluster alignment with a teacher for unsupervised domain adaptation.
\newblock In {\em ICCV}, 2019.

\bibitem{everingham2010pascal}
Mark Everingham, Luc Van~Gool, Christopher~KI Williams, John Winn, and Andrew
  Zisserman.
\newblock The pascal visual object classes (voc) challenge.
\newblock {\em IJCV}, pages 303--338, 2010.

\bibitem{fernando2013unsupervised}
Basura Fernando, Amaury Habrard, Marc Sebban, and Tinne Tuytelaars.
\newblock Unsupervised visual domain adaptation using subspace alignment.
\newblock In {\em ICCV}, pages 2960--2967, 2013.

\bibitem{ganin2015unsupervised}
Yaroslav Ganin and Victor Lempitsky.
\newblock Unsupervised domain adaptation by backpropagation.
\newblock In {\em ICML}, pages 1180--1189, 2015.

\bibitem{ganin2016domain}
Yaroslav Ganin, Evgeniya Ustinova, Hana Ajakan, Pascal Germain, Hugo
  Larochelle, Fran{\c{c}}ois Laviolette, Mario Marchand, and Victor Lempitsky.
\newblock Domain-adversarial training of neural networks.
\newblock {\em JMLR}, 17(1):2096--2030, 2016.

\bibitem{gong2012geodesic}
Boqing Gong, Yuan Shi, Fei Sha, and Kristen Grauman.
\newblock Geodesic flow kernel for unsupervised domain adaptation.
\newblock In {\em CVPR}, pages 2066--2073, 2012.

\bibitem{goodfellow2014generative}
Ian Goodfellow, Jean Pouget-Abadie, Mehdi Mirza, Bing Xu, David Warde-Farley,
  Sherjil Ozair, Aaron Courville, and Yoshua Bengio.
\newblock Generative adversarial nets.
\newblock In {\em NIPS}, pages 2672--2680, 2014.

\bibitem{he2016deep}
Kaiming He, Xiangyu Zhang, Shaoqing Ren, and Jian Sun.
\newblock Deep residual learning for image recognition.
\newblock In {\em Proceedings of the IEEE conference on computer vision and
  pattern recognition}, pages 770--778, 2016.

\bibitem{He_2019_ICCV}
Zhenwei He and Lei Zhang.
\newblock Multi-adversarial faster-rcnn for unrestricted object detection.
\newblock In {\em ICCV}, 2019.

\bibitem{hoffman2018cycada}
Judy Hoffman, Eric Tzeng, Taesung Park, Jun-Yan Zhu, Phillip Isola, Kate
  Saenko, Alexei Efros, and Trevor Darrell.
\newblock Cycada: Cycle-consistent adversarial domain adaptation.
\newblock In {\em ICML}, pages 1994--2003, 2018.

\bibitem{hu2018duplex}
Lanqing Hu, Meina Kan, Shiguang Shan, and Xilin Chen.
\newblock Duplex generative adversarial network for unsupervised domain
  adaptation.
\newblock In {\em CVPR}, pages 1498--1507, 2018.

\bibitem{inoue2018cross}
Naoto Inoue, Ryosuke Furuta, Toshihiko Yamasaki, and Kiyoharu Aizawa.
\newblock Cross-domain weakly-supervised object detection through progressive
  domain adaptation.
\newblock In {\em CVPR}, pages 5001--5009, 2018.

\bibitem{johnson2017driving}
Matthew Johnson-Roberson, Charles Barto, Rounak Mehta, Sharath~Nittur Sridhar,
  Karl Rosaen, and Ram Vasudevan.
\newblock Driving in the matrix: Can virtual worlds replace human-generated
  annotations for real world tasks?
\newblock In {\em ICRA}, pages 746--753, 2017.

\bibitem{Kang_2019_CVPR}
Guoliang Kang, Lu Jiang, Yi Yang, and Alexander~G. Hauptmann.
\newblock Contrastive adaptation network for unsupervised domain adaptation.
\newblock In {\em CVPR}, 2019.

\bibitem{kar2012random}
Purushottam Kar and Harish Karnick.
\newblock Random feature maps for dot product kernels.
\newblock In {\em Artificial Intelligence and Statistics}, pages 583--591,
  2012.

\bibitem{kim2019self}
Seunghyeon Kim, Jaehoon Choi, Taekyung Kim, and Changick Kim.
\newblock Self-training and adversarial background regularization for
  unsupervised domain adaptive one-stage object detection.
\newblock In {\em ICCV}, pages 6092--6101, 2019.

\bibitem{kim2019diversify}
Taekyung Kim, Minki Jeong, Seunghyeon Kim, Seokeon Choi, and Changick Kim.
\newblock Diversify and match: A domain adaptive representation learning
  paradigm for object detection.
\newblock In {\em CVPR}, pages 12456--12465, 2019.

\bibitem{kumar2018co}
Abhishek Kumar, Prasanna Sattigeri, Kahini Wadhawan, Leonid Karlinsky, Rogerio
  Feris, Bill Freeman, and Gregory Wornell.
\newblock Co-regularized alignment for unsupervised domain adaptation.
\newblock In {\em NIPS}, pages 9345--9356, 2018.

\bibitem{Li_2019_CVPR}
Yunsheng Li, Lu Yuan, and Nuno Vasconcelos.
\newblock Bidirectional learning for domain adaptation of semantic
  segmentation.
\newblock In {\em CVPR}, 2019.

\bibitem{lian2019constructing}
Qing Lian, Fengmao Lv, Lixin Duan, and Boqing Gong.
\newblock Constructing self-motivated pyramid curriculums for cross-domain
  semantic segmentation: A non-adversarial approach.
\newblock In {\em ICCV}, 2019.

\bibitem{lin2017focal}
Tsung-Yi Lin, Priya Goyal, Ross Girshick, Kaiming He, and Piotr Doll{\'a}r.
\newblock Focal loss for dense object detection.
\newblock In {\em ICCV}, pages 2980--2988, 2017.

\bibitem{liu2017unsupervised}
Ming-Yu Liu, Thomas Breuel, and Jan Kautz.
\newblock Unsupervised image-to-image translation networks.
\newblock In {\em NIPS}, pages 700--708, 2017.

\bibitem{liu2016ssd}
Wei Liu, Dragomir Anguelov, Dumitru Erhan, Christian Szegedy, Scott Reed,
  Cheng-Yang Fu, and Alexander~C Berg.
\newblock Ssd: Single shot multibox detector.
\newblock In {\em ECCV}, pages 21--37, 2016.

\bibitem{long2015learning}
Mingsheng Long, Yue Cao, Jianmin Wang, and Michael Jordan.
\newblock Learning transferable features with deep adaptation networks.
\newblock In {\em ICML}, pages 97--105, 2015.

\bibitem{long2018conditional}
Mingsheng Long, Zhangjie Cao, Jianmin Wang, and Michael~I Jordan.
\newblock Conditional adversarial domain adaptation.
\newblock In {\em NIPS}, pages 1640--1650, 2018.

\bibitem{long2017deep}
Mingsheng Long, Han Zhu, Jianmin Wang, and Michael~I Jordan.
\newblock Deep transfer learning with joint adaptation networks.
\newblock In {\em ICML}, pages 2208--2217, 2017.

\bibitem{pan2010survey}
Sinno~Jialin Pan and Qiang Yang.
\newblock A survey on transfer learning.
\newblock {\em IEEE Transactions on knowledge and data engineering},
  22(10):1345--1359, 2010.

\bibitem{Pan_2019_CVPR}
Yingwei Pan, Ting Yao, Yehao Li, Yu Wang, Chong-Wah Ngo, and Tao Mei.
\newblock Transferrable prototypical networks for unsupervised domain
  adaptation.
\newblock In {\em CVPR}, 2019.

\bibitem{pei2018multi}
Zhongyi Pei, Zhangjie Cao, Mingsheng Long, and Jianmin Wang.
\newblock Multi-adversarial domain adaptation.
\newblock In {\em AAAI}, 2018.

\bibitem{peng2018megdet}
Chao Peng, Tete Xiao, Zeming Li, Yuning Jiang, Xiangyu Zhang, Kai Jia, Gang Yu,
  and Jian Sun.
\newblock Megdet: A large mini-batch object detector.
\newblock In {\em CVPR}, pages 6181--6189, 2018.

\bibitem{peng2019moment}
Xingchao Peng, Qinxun Bai, Xide Xia, Zijun Huang, Kate Saenko, and Bo Wang.
\newblock Moment matching for multi-source domain adaptation.
\newblock In {\em ICCV}, 2019.

\bibitem{redmon2017yolo9000}
Joseph Redmon and Ali Farhadi.
\newblock Yolo9000: better, faster, stronger.
\newblock In {\em CVPR}, pages 7263--7271, 2017.

\bibitem{ren2015faster}
Shaoqing Ren, Kaiming He, Ross Girshick, and Jian Sun.
\newblock Faster r-cnn: Towards real-time object detection with region proposal
  networks.
\newblock In {\em NIPS}, pages 91--99, 2015.

\bibitem{russo2018source}
Paolo Russo, Fabio~M Carlucci, Tatiana Tommasi, and Barbara Caputo.
\newblock From source to target and back: symmetric bi-directional adaptive
  gan.
\newblock In {\em CVPR}, pages 8099--8108, 2018.

\bibitem{saito2019strong}
Kuniaki Saito, Yoshitaka Ushiku, Tatsuya Harada, and Kate Saenko.
\newblock Strong-weak distribution alignment for adaptive object detection.
\newblock In {\em CVPR}, pages 6956--6965, 2019.

\bibitem{saito2018maximum}
Kuniaki Saito, Kohei Watanabe, Yoshitaka Ushiku, and Tatsuya Harada.
\newblock Maximum classifier discrepancy for unsupervised domain adaptation.
\newblock In {\em CVPR}, pages 3723--3732, 2018.

\bibitem{shen2018wasserstein}
Jian Shen, Yanru Qu, Weinan Zhang, and Yong Yu.
\newblock Wasserstein distance guided representation learning for domain
  adaptation.
\newblock In {\em AAAI}, 2018.

\bibitem{shu2018dirt}
Rui Shu, Hung~H Bui, Hirokazu Narui, and Stefano Ermon.
\newblock A dirt-t approach to unsupervised domain adaptation.
\newblock In {\em ICLR}, 2018.

\bibitem{simonyan2014very}
Karen Simonyan and Andrew Zisserman.
\newblock Very deep convolutional networks for large-scale image recognition.
\newblock {\em arXiv preprint arXiv:1409.1556}, 2014.

\bibitem{sun2016deep}
Baochen Sun and Kate Saenko.
\newblock Deep coral: Correlation alignment for deep domain adaptation.
\newblock In {\em ECCV}, pages 443--450. Springer, 2016.

\bibitem{torralba2011unbiased}
Antonio Torralba and Alexei~A Efros.
\newblock Unbiased look at dataset bias.
\newblock In {\em Computer Vision and Pattern Recognition (CVPR), 2011 IEEE
  Conference on}, pages 1521--1528. IEEE, 2011.

\bibitem{Tran_2019_CVPR}
Luan Tran, Kihyuk Sohn, Xiang Yu, Xiaoming Liu, and Manmohan Chandraker.
\newblock Gotta adapt 'em all: Joint pixel and feature-level domain adaptation
  for recognition in the wild.
\newblock In {\em CVPR}, 2019.

\bibitem{tsai2018learning}
Yi-Hsuan Tsai, Wei-Chih Hung, Samuel Schulter, Kihyuk Sohn, Ming-Hsuan Yang,
  and Manmohan Chandraker.
\newblock Learning to adapt structured output space for semantic segmentation.
\newblock In {\em CVPR}, pages 7472--7481, 2018.

\bibitem{tsipras2019robustness}
Dimitris Tsipras, Shibani Santurkar, Logan Engstrom, Alexander Turner, and
  Aleksander Madry.
\newblock Robustness may be at odds with accuracy.
\newblock In {\em ICLR}, 2019.

\bibitem{tzeng2017adversarial}
Eric Tzeng, Judy Hoffman, Kate Saenko, and Trevor Darrell.
\newblock Adversarial discriminative domain adaptation.
\newblock In {\em CVPR}, 2017.

\bibitem{tzeng2014deep}
Eric Tzeng, Judy Hoffman, Ning Zhang, Kate Saenko, and Trevor Darrell.
\newblock Deep domain confusion: Maximizing for domain invariance.
\newblock {\em arXiv preprint arXiv:1412.3474}, 2014.

\bibitem{Wang_2019_CVPR}
Zirui Wang, Zihang Dai, Barnabas Poczos, and Jaime Carbonell.
\newblock Characterizing and avoiding negative transfer.
\newblock In {\em CVPR}, 2019.

\bibitem{xie2018learning}
Shaoan Xie, Zibin Zheng, Liang Chen, and Chuan Chen.
\newblock Learning semantic representations for unsupervised domain adaptation.
\newblock In {\em ICML}, pages 5419--5428, 2018.

\bibitem{yosinski2014transferable}
Jason Yosinski, Jeff Clune, Yoshua Bengio, and Hod Lipson.
\newblock How transferable are features in deep neural networks?
\newblock In {\em NIPS}, pages 3320--3328, 2014.

\bibitem{zellinger2017central}
Werner Zellinger, Thomas Grubinger, Edwin Lughofer, Thomas Natschl{\"a}ger, and
  Susanne Saminger-Platz.
\newblock Central moment discrepancy (cmd) for domain-invariant representation
  learning.
\newblock In {\em ICLR}, 2017.

\bibitem{zhu2017unpaired}
Jun-Yan Zhu, Taesung Park, Phillip Isola, and Alexei~A Efros.
\newblock Unpaired image-to-image translation using cycle-consistent
  adversarial networks.
\newblock In {\em ICCV}, pages 2223--2232, 2017.

\bibitem{zhu2019adapting}
Xinge Zhu, Jiangmiao Pang, Ceyuan Yang, Jianping Shi, and Dahua Lin.
\newblock Adapting object detectors via selective cross-domain alignment.
\newblock In {\em CVPR}, pages 687--696, 2019.

\bibitem{zhu2018penalizing}
Xinge Zhu, Hui Zhou, Ceyuan Yang, Jianping Shi, and Dahua Lin.
\newblock Penalizing top performers: Conservative loss for semantic
  segmentation adaptation.
\newblock In {\em Proceedings of the European Conference on Computer Vision
  (ECCV)}, pages 568--583, 2018.

\bibitem{zou2018unsupervised}
Yang Zou, Zhiding Yu, BVK Vijaya~Kumar, and Jinsong Wang.
\newblock Unsupervised domain adaptation for semantic segmentation via
  class-balanced self-training.
\newblock In {\em ECCV}, pages 289--305, 2018.

\end{thebibliography}
}

\end{document}